\documentclass[10pt,journal,compsoc]{IEEEtran}
\ifCLASSOPTIONcompsoc
  \usepackage[nocompress]{cite}
\else
  \usepackage{cite}
\fi
\ifCLASSINFOpdf
  \usepackage[pdftex]{graphicx}
\else
\fi
\usepackage{amsmath}
\usepackage{algorithmic}

\usepackage{array}

\ifCLASSOPTIONcompsoc
  \usepackage[caption=false,font=footnotesize,labelfont=sf,textfont=sf]{subfig}
\else
  \usepackage[caption=false,font=footnotesize]{subfig}
\fi

\usepackage{stfloats}
\usepackage{url}

\usepackage{amsmath}
\usepackage{graphicx}
\usepackage[colorinlistoftodos]{todonotes}
\usepackage[colorlinks=true, allcolors=blue]{hyperref}

\usepackage{comment}

\usepackage{color}
\definecolor{Orange}{rgb}{1,0.5,0}
\definecolor{Red}{rgb}{1,0,0}
\definecolor{Blue}{rgb}{0,0,1}

\newcommand{\BL}[1]{\textsf{\textbf{\textcolor{Red}{\footnotesize [BL: #1]}}}}

\usepackage{bm}
\usepackage{amsfonts} 
\usepackage{amssymb} 
\usepackage{stmaryrd} 
\DeclareMathOperator*{\argmax}{arg\,max}
\DeclareMathOperator*{\argmin}{arg\,min}

\usepackage{booktabs}
\usepackage{longtable}
\usepackage{array}
\usepackage{multirow}

\usepackage[linesnumbered,ruled,vlined]{algorithm2e}

\usepackage{enumitem}
\setlist[itemize]{leftmargin=*}

\begin{document}
%
\title{Multi-Label Sampling \\ based on Local Label Imbalance}
%
%
%
%

\author{Bin~Liu, Konstantinos~Blekas, and~Grigorios~Tsoumakas
\IEEEcompsocitemizethanks{\IEEEcompsocthanksitem B. Liu and G. Tsoumakas are with the School of Informatics, Aristotle University of Thessaloniki, Thessaloniki 54124, Greece. E-mail: \{binliu, greg\}@csd.auth.gr. \protect\\
\IEEEcompsocthanksitem {K. Blekas is with the Department of Computer Science and Engineering, University of Ioannina, Ioannina 45110, Greece. E-mail: kblekas@cs.uoi.gr}.}
}

\IEEEtitleabstractindextext{%
\begin{abstract}
Class imbalance is an inherent characteristic of multi-label data that hinders most multi-label learning methods.
One efficient and flexible strategy to deal with this problem is to employ sampling techniques before training a multi-label learning model. Although existing multi-label sampling approaches alleviate the global imbalance of multi-label datasets, it is actually the imbalance level within the local neighbourhood of minority class examples that plays a key role in performance degradation. To address this issue, we propose a novel measure to assess the local label imbalance of multi-label datasets, as well as two multi-label sampling approaches based on the local label imbalance, namely MLSOL and MLUL. By considering all informative labels, MLSOL creates more diverse and better labeled synthetic instances for difficult examples, while MLUL eliminates instances that are harmful to their local region.
Experimental results on 13 multi-label datasets demonstrate the effectiveness of the proposed measure and sampling approaches for a variety of evaluation metrics, particularly in the case of an ensemble of classifiers trained on repeated samples of the original data.
\end{abstract}

}

\maketitle

\IEEEdisplaynontitleabstractindextext

%
\IEEEpeerreviewmaketitle


%
%
%
%
\section{Introduction \label{sec:introduction}}
\IEEEPARstart{I}{n} multi-label data, each instance is associated with multiple binary output variables (labels), which allow the expression of much richer semantics compared to binary and multi-class data. The number of labels assigned to each instance is typically much smaller than the total number of output variables. In consequence, the number of instances relevant to each label is much less than the number of irrelevant ones. This gives rise to the problem of {\em class imbalance}, which has been recently recognized as a key challenge in multi-label learning\cite{Charte2015,Charte2015a,Daniels2017,Liu2018MakingImbalance,Zhang2015a}.

There are two main types of methods for handling class imbalance in multi-label data: multi-label sampling and algorithm adaptation. The former reduce the imbalance level of multi-label data via adding or removing instances as a pre-processing  step\cite{Charte2015a,Charte2015,Charte2014,Charte2019}. The latter make multi-label learning approaches resilient to class imbalance directly\cite{Daniels2017,Liu2018MakingImbalance,Zhang2015a}. This work focuses on multi-label sampling methods, which can be coupled with any multi-label learning algorithm and are therefore more flexible.


A key challenge for multi-label sampling methods, which has not yet been properly addressed, is how to deal with the co-occurrence of multiple labels, which have varying frequencies, in the same training example.  MLROS\cite{Charte2015} and MLSMOTE\cite{Charte2015a} relieve the imbalance level for each single minority (less frequent) label via duplicating or generating instances, but this can lead to other labels suffering more severe imbalance. Similarly, MLRUS\cite{Charte2015} reduces the imbalance by focusing on the majority (higher frequent) label separately via removing instances, but this may increase the imbalance level of other labels as well. MLeNN\cite{Charte2014} removes examples associated with majority labels only, yet it is unable to process complex examples having minority and majority labels simultaneously. REMEDIAL\cite{Charte2019} divides a complex example into two easier examples, of which one is associated with minority labels and another with majority labels. However, REMEDIAL brings in additional noise in the dataset, because the pair of new examples have identical features but different labels.


In essence, all the above sampling approaches for multi-label data focus on class imbalance at the {\em global} scale of the whole dataset. However, previous studies of binary and multi-class data have found that the main reason for the difficulty of a classifier to recognize the minority class is the distribution of class values in the {\em local} neighbourhood of the minority examples\cite{napierala2016types,Saez2016}.
We hypothesize that in a similar vein, the \textit{local distribution of the labels} is more important than the global imbalance level of each label to determine the hardness of a multi-label dataset to be learned.

\begin{figure}
\begin{center}
\includegraphics[width=0.48\textwidth]{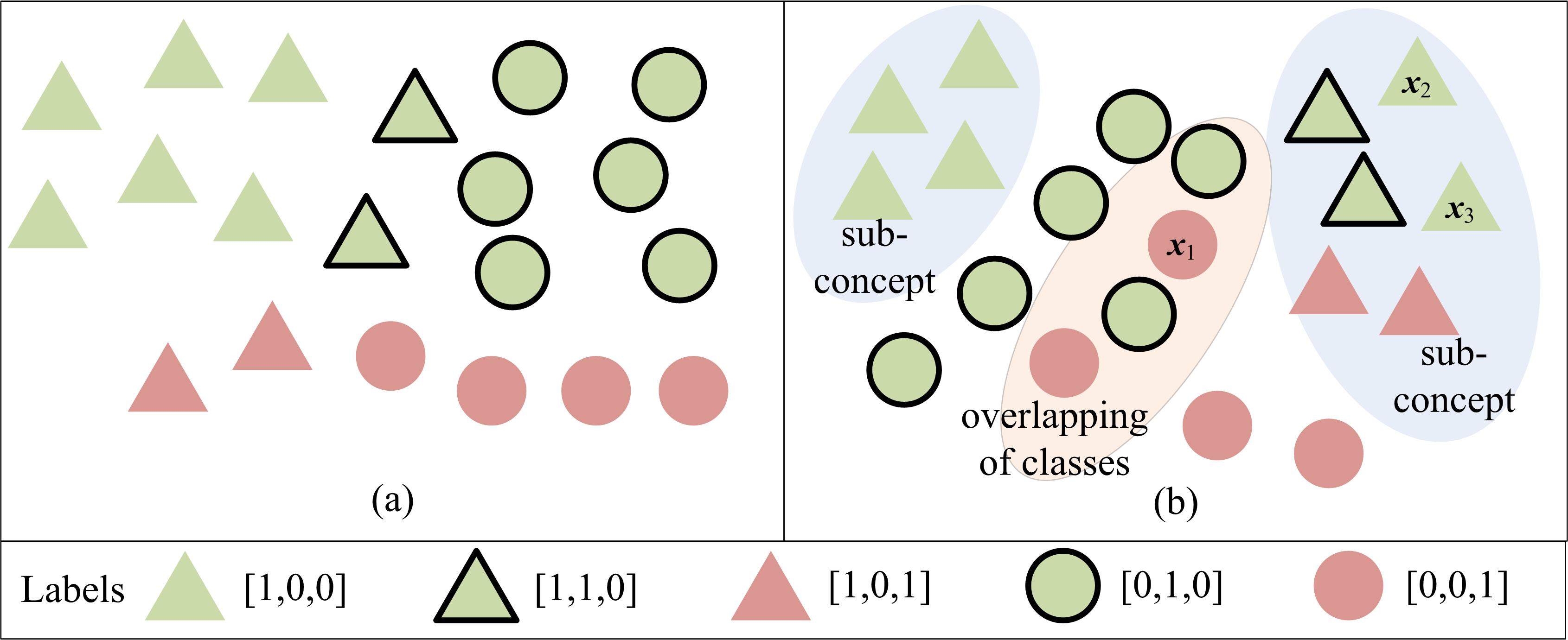}
\caption{Two 2-dimensional multi-label datasets (a) and (b) concerning points in a plane characterized by three labels, namely the shape of the points (triangles, circles), the border of the points (solid, none) and the color of the points (green, red). In the bottom we see the five different label combinations that exist in datasets. The two datasets have same global imbalance level per label because the number of relevant instances for the three labels is 10,8,6 respectively in both (a) and (b). While, the local label distribution of (b) is more complex than (a) due to the appearance of sub-concept and overlapping of classes.}
\label{fig:ExampleDif}
\end{center}
\end{figure}

Fig.\ref{fig:ExampleDif} shows an example of two multi-label datasets with the same global level of label imbalance, but different local label distribution. Indeed, dataset (b) appears much more challenging than (a) due to its more complex local label distribution caused by the presence of sub-concepts for the triangle and border class, as well as the overlap of the color classes along with the border style classes.

Starting from our hypothesis, we first present a measure for assessing the local imbalance level of a multi-label dataset based on the local distribution of the labels. We then propose two twin multi-label sampling methods that take the local imbalance of the labels into account, namely Multi-Label Synthetic Oversampling based on Local label imbalance (MLSOL) and Multi-Label Undersampling based on Local label imbalance (MLUL). MLSOL creates new instances near difficult to learn examples, by using the local label imbalance within the seed instance selection and synthetic instance generation processes. MLUL eliminates difficult to learn examples as evaluated by both local label imbalance and the influence of the example on its reversed $k$ nearest neighbours (R$k$NN). MLSOL and MLUL take all labels in one instance appropriately into account, by considering the influence of all informative labels. Finally, we embed MLSOL and MLUL, as well as other multi-label sampling methods, within a simple but flexible ensemble framework to further improve their performance and robustness.
Experimental results on 13 multi-label datasets illustrate the validity of the proposed measure to assess the difficulty of multi-label dataset and demonstrate the effectiveness of the proposed sampling approaches. 


This paper extends our previous work\cite{Liu2019Synthetic} in the following aspects:
\begin{itemize}
\item We discuss existing measures to evaluate the imbalance level of multi-label data and propose a new measure based on local label imbalance.
\item We propose MLUL, which removes harmful instances via combining the instance's difficulty and impact on its R$k$NN.
\item We empirically validate the effectiveness of the proposed measure, investigate the reasons why our sampling approaches benefit more from the ensemble framework, and examine the influence of different parameter settings on the proposed methods for easy and difficult dataset respectively.
\end{itemize}


The remainder of this paper is organized as follows. Section 2 offers a brief review of previous work. In Section 3, we introduce the new measure to assess the local imbalance level of multi-label data, propose the two sampling approaches based on local label distribution, and present the ensemble framework that can be coupled with multi-label sampling methods. Then, the experimental results along with their discussion is included in Section 4. Finally, the main conclusions of this work are given in Section 5.


\section{Related Work \label{sec:relatedWork}}
We first briefly introduce the multi-label learning problem. Then, we present existing measures for assessing the imbalance level of multi-label datasets. Next, we review methods that deal with the class imbalance issue in multi-label learning. Lastly, we discuss sampling approaches for dealing with class imbalance in binary and multi-class classification.

\subsection{Multi-Label Learning}

Let $\mathcal{X}=\mathbb{R}^d$ be a $d$-dimensional input feature space, $L=\{l_1,l_2,...,l_q\}$ a label set containing $q$ labels and $\mathcal{Y}=\{0,1\}^q$ a $q$-dimensional label space. Let $D=\{(\bm{x}_i,\bm{y}_i) \bigm\lvert 1 \leqslant i \leqslant n \}$ be a multi-label training set containing $n$ instances. Each instance $(\bm{x}_i,\bm{y}_i)$ consists of a feature vector $\bm{x}_i \in \mathcal{X}$ and a label vector $\bm{y}_i \in \mathcal{Y}$, where $y_{ij}$ is the $j$-th element of $\bm{y}_i$ and $y_{ij}=1 (0)$ denotes that $l_j$ is (not) associated with the $i$-th instance. The goal of multi-label learning is to learn a mapping function $h:\mathcal{X} \to \{0,1\}^q$ and (or) $f: \mathcal{X} \to \mathbb{R}^q$ that given an unseen instance $\bm{x} \in \mathcal{X}$, outputs a label vector $\hat{\bm{y}}$ with the predicted labels and (or) real-valued vector $\hat{\bm{fy}}$ with the corresponding relevance degrees to $\bm{x}$ respectively.

The main goal of most multi-label learning methods is to exploit the correlations among labels in order to improve prediction accuracy~\cite{Zhu2018a,Akbarnejad2019,Li2019}. Multi-label learning methods are divided into three families based on the order of label correlations that they consider, namely first-order, second-order and high-order~\cite{Zhang2014}. BR~\cite{Boutell2004} and MLkNN~\cite{zhang2007ml} are first-order strategies that treat all labels independently and ignore label dependencies.  CLR~\cite{Furnkranz2008} is a representative second-order approach that considers the correlation among pairs of labels via transforming the multi-label learning problem into several pair-wise label ranking problems. RAkEL~\cite{tsoumakas2011random} and ECC~\cite{Read2011} are two methods that exploit high order label correlations by treating label subsets as classes and by embedding labels in chain models respectively.

\subsection{Measuring the Imbalance of Multi-Label Data}
The imbalance level of a single-label (binary, multi-class) dataset is typically measured by the imbalance ratio, which is computed as the proportion of the number of majority class instances to the number of minority class instances\cite{He2009}.

In multi-label learning, two measures that evaluate the imbalance of a particular label are $IRLbl$~\cite{Charte2015} and $ImR$~\cite{Zhang2015a,Liu2018MakingImbalance}. Let $n_j^b=|\{(\bm{x}_i,y_{ij}) \bigm\lvert y_{ij}=b, 1 \leqslant i \leqslant n\}|$ be the number of instances whose $j$-th label value is equal to $b \in \{0,1\}$.
Let $G_j=\argmax_{b \in \{0,1\}}n_j^b$ and $g_j=\argmin_{b \in \{0,1\}}n_j^b$ denote the majority and minority class of $l_j$ respectively. $IRLbl$ and $ImR$ are then formally defined as follows:

\begin{equation}
IRLbl_j=\frac{1}{n_j^1}\max\limits_{k=1,...,q}\{n_k^1 \} ,\  j=1,2,...,q,
\label{eq:IRLbl}
\end{equation}
\begin{equation}
ImR_j = \left. n_j^{G_j} \middle/ n_j^{g_j} \right.,\  j=1,2,...,q
\label{eq:ImR}
\end{equation}


By considering the average and the coefficient of variation of $IRLbl$ and $ImR$ across all labels, four measures of the imbalance of multi-label data have been proposed~\cite{Charte2015,Liu2018MakingImbalance}:
\begin{equation}
MeanIR=\frac{1}{|q|}\sum\limits_{j=1}^{|q|}IRLbl_j
\label{eq:MeanIR}
\end{equation}
\begin{equation}
MeanImR=\frac{1}{|q|}\sum\limits_{j=1}^{q} ImR_j
\label{eq:MeanImR}
\end{equation}
\begin{equation}
CVIR=\frac{1}{MeanIR}\sqrt{\sum\limits_{j=1}^{q}\frac{\left(IRLbl_j - MeanIR\right)^2}{q-1}}
\label{eq:CVIR}
\end{equation}
\begin{equation}
CVImR=\frac{1}{MeanImR}\sqrt{\sum\limits_{j=1}^{q}\frac{\left(ImR_j - MeanImR\right)^2}{q-1}}
\label{eq:CVImR}
\end{equation}

Labels whose $IRLbl$ is larger (less) than $MeanIR$ are called \textit{majority} (\textit{minority}) in \textit{label}\cite{Charte2015}. The coefficient of variation examines whether all labels suffer from a similar or different level of imbalance. For all of the above measures, the higher the value, the more imbalanced the dataset.

When minority class is "1", which is the typical situation for multi-label datasets, $IRLbl$ is linearly correlated with $ImR$:
\begin{equation}
\begin{aligned}
 ImR_j=n*IRLbl_j/ \max\limits_{k=1,...,q}\{n_k^1\}-1 \\
\end{aligned}
\label{eq:IRLbl_ImR}
\end{equation}
Based on Eq.\eqref{eq:IRLbl_ImR}, the relation between the $IRLbl$ and $ImR$ based measures are:
\begin{equation}
\begin{aligned}
 & MeanImR=n*MeanIR/\max\limits_{j=1,...,q}\{n_j^1\}-1 \\
 & CVImR= \frac{MeanIR*CVIR}{(MeanIR -\max\limits_{j=1,...,q}\{n_j^1\}/n)}    \\
\end{aligned}
\end{equation}
However, if the minority class is "0", then the two groups of measures are different because $IRLbl_j$ uses $|D^1_j|$ as numerator while the denominator in $ImR_j$ is $|D^0_j|$.

A recent measure of the imbalance in a multi-label data set that takes into account the occurrence of frequent and rare labels is $SCUMBLE$~\cite{Charte2019}. It is computed based on the Atkinson index and $IRLbl$  as follows:
\begin{equation}
SCUMBLE=\frac{1}{n}\sum_{i=1}^{n}SCUins_i
\label{eq:SCUMBLE}
\end{equation}
\begin{equation}
SCUins_i=1-\frac{\sum_{j=1}^{q} y_{ij} \left(\prod_{j=1}^{q} \left( IRLbl_{j} \right) ^{y_{ij}}\right)^\frac{1}{\sum_{j=1}^{q} y_{ij}}}{ \sum_{j=1}^{q} y_{ij}IRLbl_j}
\label{eq:SCUMBLEins}
\end{equation}

The range of $SCUMBLE$ is between 0 and 1, with higher values indicating more inconsistent frequencies of labels in the examples.

\subsection{Handling the Imbalance of Multi-Label Data}

Existing methods for dealing with the class imbalance issue in multi-label data can be divided in two groups: multi-label sampling and algorithm adaptation.

\subsubsection{Sampling Methods}

Multi-label sampling methods relieve the global imbalance level of the whole dataset by manipulating the training instances in a pre-processing step. They are independent of the particular multi-label learning algorithm that will be subsequently applied to the dataset.

Multi-label undersampling methods delete instances to reduce the imbalance of the dataset. LP-RUS interprets each labelset (i.e. particular combination of label values) as class identifier and removes instances assigned with the most frequent labelset~\cite{charte2013}. Instead of considering the whole labelset, MLRUS alleviates the imbalance of the dataset in the individual label aspect via omitting instances with majority labels randomly~\cite{Charte2015}. MLeNN employs an Edited Nearest Neighbor (ENN) based strategy to heuristically eliminate instances only assigned with majority labels and have similar labelset with their neighbors ~\cite{Charte2014}.

To achieve the balanced label distribution, multi-label oversampling approaches add instances to the dataset. LP-ROS, as a twin method of LP-RUS, replicates instances whose labelset appears the fewest times~\cite{charte2013}. Similar to MLRUS, MLROS increases the frequency of minority labels via replicating instances relevant to minority labels to relieve the imbalance in view of individual labels~\cite{Charte2015}. To reduce the risk of overfitting caused by copying instances, MLSMOTE randomly selects an instance containing minority labels, along with its neighbors, to generate synthetic instances. These instances are associated with the labels that appear in more than half of the seed instance and its neighbors~\cite{Charte2015a}.

REMEDIAL tackles the co-ocurrence of labels with different imbalance level in one instance, of which the level is assessed by $SCUMBLE$, by decomposing the sophisticated instance into two simpler examples, but may introduce extra confusions into the learning task, i.e. there are several pairs of instances with same features and different labels~\cite{Charte2019}. REMEDIAL can be used either as standalone sampling method or the prior part of another sampling technique. For example, RHwRSMT combines REMEDIAL with MLSMOTE~\cite{Charte2019a}.

\subsubsection{Algorithm Adaptation Methods}

Different from sampling methods, algorithm adaptation methods focus on the multi-label learning algorithm handling the class imbalance problem directly. One kind of methods deal with the imbalance issue of multi-label learning via transforming the multi-label dataset to several binary/multi-class classification problems. COCOA converts the original multi-label dataset to one binary dataset and several multi-class datasets for each label, and builds imbalance classifiers with the assistance of sampling for each dataset~\cite{Zhang2015a}. SOSHF transforms the multi-label learning task to an imbalanced single label classification assignment via cost-sensitive clustering, and the new task is addressed by oblique structured Hellinger decision trees~\cite{Daniels2017}.

Another branch of approaches aims to modify current multi-label learning methods to handle the class imbalance problem. ECCRU3 makes ECC resilient to class imbalance by coupling it with undersampling and improving the exploitation of majority examples~\cite{Liu2018MakingImbalance}. Apart from ECCRU3, modified models based on neural networks~\cite{Tepvorachai2008,Li2013,Sozykin2017}, SVM\cite{Cao2016}, hypernetwork~\cite{Sun2017} and BR~\cite{Chen2006,Dendamrongvit2010UndersamplingDomains,Tahir2012,Wan2017} have been proposed as well.

Furthermore, other strategies, such as representation learning~\cite{LiWang16}, constrained submodular minimization~\cite{Wu:2016} and balanced pseudo-label~\cite{Zeng2014} have been utilized to address the imbalance obstacle of multi-label learning as well.

\subsection{Sampling for Binary and Multi-Class Imbalance}
Sampling approaches are widely used to deal with the class imbalance issue in traditional binary and multi-class classification~\cite{He2009}.
Undersampling approaches remove majority instances that are near to minority instances (in the class boundary region)~\cite{Wilson1972}.
SMOTE is a well known synthetic oversampling approach that creates new minority class instances based on a randomly selected minority class example and it's nearest neighbours from the same class~\cite{Chawla2002}. Several extensions of SMOTE, such as Borderline SMOTE~\cite{Han2005}, Safe-level SMOTE~\cite{Bunkhumpornpat2009}, ADASYN~\cite{He2008} and MWMOTE~\cite{Barua2014}, employ more effective strategies to generate synthetic instances around unsafe or important minority instances.

Recently, several advanced sampling approaches that depend on Mahalanobis distance~\cite{Abdi2016,Yang2017}, kernel based adaptive subspace~\cite{Lin2018a} and entropy-based imbalance degree~\cite{li2019entropy} have been proposed to address the class imbalance issue in various respects.
Furthermore, the combination of sampling with ensemble methods has been found to improve its performance and robustness~\cite{Galar2012,Tang2017}.

\section{Our Contributions} \label{sec:approaches}

We first present a new measure for evaluating the local imbalance of a multi-label dataset, by considering the labels of the instances in the neighbourhood of each instance. Then, we propose two new multi-label sampling approaches based on local label distribution. Subsequently, we introduce a simple but flexible framework for ensembling multi-label sampling approaches. Lastly, we analyze the computational complexity of the proposed approaches.

\subsection{Local Imbalance of Multi-Label Data}
As we illustrated in Figure \ref{fig:ExampleDif}, the local label distribution rather than the global imbalance level is what makes a multi-label dataset challenging to learn. However, all existing measures for assessing the imbalance of multi-label datasets are based on the global imbalance level and ignore local information. Inspired by~\cite{napierala2016types}, we propose a measure that gauges the local imbalance of a multi-label dataset via considering the local label distribution of all instances.

The local imbalance of an instance can be measured by the proportion of opposite class values in its local neighborhood.
Specifically, for each instance $\bm{x}_i$ we first retrieve its $k$ nearest neighbours ${\cal N}^k_i$ according to a distance function, such as the Euclidean distance. Then, for each label $l_j$ we compute the proportion of neighbours having an opposite class with respect to the class of $\bm{x}_i$, as shown in Eq.\eqref{eq:Cij},
\begin{equation}
C_{ij}=\frac{1}{k}\sum_{\bm{x}_m \in {\cal N}^k_i} \llbracket y_{mj} \neq y_{ij}\rrbracket
\label{eq:Cij}
\end{equation}
where $\llbracket \pi \rrbracket$ is the indicator function that returns 1 if $\pi$ is true and 0 otherwise. The larger $C_{ij}$ is, the more imbalanced $l_j$ is in the local area of $\bm{x}_i$.
Furthermore, $C_{ij}$ not only represents the local imbalance of $\bm{x}_i$ for $l_j$, but also measures the difficulty of predicting $l_j$ correctly for $\bm{x}_i$. The value of $C_{ij}$ is in $[0,1]$, with values close to 0 (1) indicating a safe (hostile) neighborhood of similarly (oppositely) labelled examples. A value of $C_{ij}=1$ can further be viewed as a hint that $x_i$ is an outlier in this neighborhood with respect to $l_j$.
We define a matrix $\bm{C} \in \mathbb{R}^{n \times q}$ to store the local imbalance of all instances for each label.

We define the local imbalance of the whole dataset, $LImb$, as the average of $C_{ij}$ for the minority class of all instances and labels:
\begin{equation}
LImb= \frac{1}{q}\sum_{j=1}^q \frac{\sum_{i=1}^{n} C_{ij}\llbracket y_{ij}=g_j \rrbracket }{ \sum_{i=1}^{n} \llbracket y_{ij}=g_j \rrbracket}
\label{eq:LImb_D}
\end{equation}
The larger the $LImb$, the more difficult the dataset to be learned.

\subsection{Oversampling with MLSOL}
We propose a new multi-label oversampling approach, Multi-Label Synthetic Oversampling based on Local label imbalance (MLSOL), that generates synthetic instances near those instances that are suffering high local imbalance.
MLSOL combines the local imbalance of informative labels to pick up difficult seed instances. It assigns appropriate labels to the created synthetic instances so as to improve the frequency of difficult labels, without introducing noise for easy labels.

Firstly, we define some important variables based on the local imbalance matrix $\bm{C}$.
To evaluate the hardness of instance $\bm{x}_i$, we define its weight as $w_i$, which characterizes the difficulty in correctly predicting the minority class values of this example by aggregating its $C_{ij}$ for all labels. An initial straightforward way to do this is to simply sum these values for labels where the instance contains the minority class: 
\begin{equation}
w_{i}=\sum_{j=1}^{q}C_{ij}\llbracket y_{ij} = g_j\rrbracket
\label{eq:wi1}
\end{equation}
However, there are two issues with Eq.\eqref{eq:wi1}. The first one is that we have also considered the outliers.
The second issue is that the global level 
of class imbalance of each label is not taken into account in this aggregation. The fewer the number of minority class samples, the higher the difficulty of correctly classifying the corresponding minority class. In contrast, Eq.\eqref{eq:wi1} treats all labels equally.
To address these two issues, we define a new matrix $\bm{S}$ that takes both global and local imbalance into account and ignores the impact of outliers:
\begin{equation}
S_{ij}=\left \{
\begin{aligned}
& \frac{C_{ij}\llbracket y_{ij} = g_j \wedge C_{ij} < 1\rrbracket}{\sum_{i'=1}^{n}C_{i'j}\llbracket y_{i'j} = g_j\wedge C_{i'j} < 1\rrbracket}, \\ & \quad\quad\quad  \text{if} \ \llbracket y_{i'j} = g_j \wedge C_{i'j} < 1\rrbracket=1  \\
& -1, \quad \text{otherwise}
\end{aligned}
\right.
\label{eq:Sij}
\end{equation}

Adding the $C_{ij} < 1$ term to the indicator functions leads to omitting the influence of outliers. We consider that $l_j$ is the \textit{informative label} of $\bm{x}_i$ if $\llbracket y_{ij} = g_j \wedge C_{ij} < 1\rrbracket=1$ ($\bm{x}_i$ is an non-outlier minority class example for $l_j$). $S_{ij} \neq -1$ only if $l_j$ is the informative label of $\bm{x}_i$. Furthermore, the values of informative labels for all instances in $\bm{S}$ are normalized so that they sum to 1 per label, 
by dividing with the sum of the values of all non-outlier minority examples of that label. This increases the relative importance of the weights of labels with fewer samples. Finally, we arrive at the following proposed aggregation:
\begin{equation}
w_{i}=\sum_{j=1}^{q}S_{ij}\llbracket S_{ij} \neq -1\rrbracket
\label{eq:wi2}
\end{equation}
where $\llbracket S_{ij} \neq -1\rrbracket$ is equivalent to $\llbracket y_{ij} = g_j \wedge C_{ij} < 1\rrbracket$. The Eq.\eqref{eq:wi2} combines all local imbalance of informative labels for $\bm{x}_i$.
Weights of all instances are stored in $\bm{w}\in \mathbb{R}^n$.

Furthermore, we introduce the definition of the type of each instance-label pair, which would be utilized to determine the appropriate labels assigned to new instances that we will create.
Following~\cite{napierala2016types}, we discretize the range $[0, 1]$ of $C_{ij}$ to define four types of minority class instances, namely safe ($SF$), borderline ($BD$), rare ($RR$) and outlier ($OT$), according to their local imbalance 
:
\begin{itemize}
    \item $SF$ :  $0 \leqslant C_{ij} < 0.3$. Safe instances are located in the region overwhelmed by minority examples.
    \item $BD$ :  $0.3 \leqslant C_{ij} < 0.7$. Borderline instances are located in the decision boundary between minority and majority classes.
    \item $RR$ :  $0.7 \leqslant C_{ij} < 1$. We further consider only those instances whose minority class neighbours are of type $RR$ or $OT$. Otherwise there are some $SF$ or $BD$ examples in the proximity, which suggests that they should be considered as $BD$. Rare instances, accompanied by isolated pairs or triples of minority class examples, are located in the majority class area and distant from the decision boundary.
    \item $OT$ :  $ C_{ij}=1$. Outliers are surrounded by majority examples.
\end{itemize}
For completeness we add the value $MJ$ in order to associate the majority class case. Let $\bm{T} \in \{SF,BD,RR,OT,MJ\}^{n \times q}$ be the type matrix and $T_{ij}$ be the type of $y_{ij}$. 

The pseudo-code of MLSOL is shown in Algorithm \ref{al:MLSOL}. Firstly, the auxiliary variables defined previously, as the weight vector $\bm{w}$ and type matrix $\bm{T}$, are calculated (lines 3-6 in Algorithm \ref{al:MLSOL}). Next, the loop describes the procedure of creating new instances (lines 8-13 in Algorithm \ref{al:MLSOL}). In each iteration, a synthetic instance is generated as follows: a seed instance $(\bm{x}_s,\bm{y}_s)$ is picked at the beginning, with the probability of selection being proportional to its weight (i.e. the more difficult instance has more chance to be selected). 
Then reference instance $(\bm{x}_r,\bm{y}_r)$ is randomly chosen from the $k$ nearest neighbours of the seed instance. 
Finally, a synthetic example is generated based on the specific seed and reference instances and added into the dataset. The above iterative procedure is terminated when the expected number of new examples are created.

\begin{algorithm}[h]
 \SetKwData{Left}{left}\SetKwData{This}{this}\SetKwData{Up}{up}
 \SetKwFunction{InitTypes}{InitTypes}\SetKwFunction{CreateIns}{CreateIns}
 \SetKwInOut{Input}{input}\SetKwInOut{Output}{output}

 \Input{multi-label data set: $D$, sampling ratio: $p$, number of nearest neighbour: $k$ }
 \Output{new data set $D'$}
 $ GenNum \leftarrow |D|*p$ \tcc*[r]{number of instances to generate}
 $D' \leftarrow D$ \;
 Find the $k$NN of each instance \;
 Calculate $\bm{C}$ according to Eq.\eqref{eq:Cij} \;
 Compute $\bm{S}$ according to Eq.\eqref{eq:Sij} \;
 Compute $\bm{w}$ according to Eq.\eqref{eq:wi2} \;
 $\bm{T} \leftarrow \InitTypes(\bm{C},k)$ \tcc*[r]{Initialize the type of instances}
\While{$GenNum>0$}{
    $(\bm{x}_s,\bm{y}_s) \leftarrow$
    Select a seed instance $(\bm{x}_s,\bm{y}_s)$ from $D$ based on $\bm{w}$\;
    Choose a reference instance $(\bm{x}_r,\bm{y}_r)$ from ${\cal N}^k_s$\;
    $(\bm{x}_{c},\bm{y}_{c}) \leftarrow \CreateIns \left( (\bm{x}_s,\bm{y}_s), T_s, (\bm{x}_r,\bm{y}_r), T_r \right)$\;
    $D' \leftarrow D' \cup (\bm{x}_{c},\bm{y}_{c})$ \;
    $GenNum \leftarrow GenNum-1$ \;
}
\KwRet{$D'$} \;
 	
 \caption{MLSOL}
  \label{al:MLSOL}
\end{algorithm}

The detailed procedure of how to determine the features and labels of a synthetic instance based on the given seed instance $(\bm{x}_s,\bm{y}_s)$ and reference instance $(\bm{x}_r,\bm{y}_r)$ along with their types is shown in Algorithm \ref{al:GenerateInstance}.
The feature values of the synthetic instance $(\bm{x}_c,\bm{y}_c)$ are interpolated along the line which connects both input samples (lines 1-2 in Algorithm \ref{al:GenerateInstance}). Once $\bm{x}_c$ is confirmed, we compute the quantity $cd \in [0,1]$, which indicates whether the synthetic instance is closer to the seed ($cd < 0.5$) or closer to the reference instance ($cd > 0.5$) (lines 3-4 in Algorithm \ref{al:GenerateInstance}).

With respect to label assignment, we employ a scheme considering the labels and types of the seed and reference instances as well as the location of the synthetic instance. This scheme is able to create informative instances for locally imbalanced labels without bringing in noises for the rest of the labels.
For each label $l_j$, $y_{cj}$ is set as $y_{sj}$ (lines 6-7 in Algorithm \ref{al:GenerateInstance}) if $y_{sj}$ and $y_{rj}$ belong to the same class.
Otherwise, in the case where $y_{sj}$ is the majority class ($T_{sj}=MJ$), the seed and reference instances are exchanged to guarantee that $y_{sj}$ is always the minority class (lines 9-11 in Algorithm \ref{al:GenerateInstance}).
Then, the threshold $\theta$ for $cd$ is defined according to the type of the label in the seed instance $T_{sj}$ (lines 12-16 in Algorithm \ref{al:GenerateInstance}),
which would be responsible for identifying the instance (seed or reference) that will lend its label to the synthetic example. In the case of the first three types ($SF$, $BD$, $RR$), where the minority class (seed) example is surrounded by several majority class instances and may lead to a wrong classification decision, the cut-point of label assignment is closer to the majority class (reference) instance. Specifically,
\begin{itemize}
    \item For safe instances ($SF$), we set $\theta=0.5$ so that the label of the nearest (seed or reference) instance is assigned to the synthetic instance.
    \item For rare instances ($RR$), the threshold takes value greater than $1$, so as to ensure that the seed's class will remain minority, i.e. $y_{cj} \leftarrow y_{sj}$.
    \item For borderline instances ($BD$), we set  $\theta = \frac{1+0.5}{2} = 0.75$ that is the midpoint between two previous cases.
    \item Finally, for obtaining smoother decision boundary in class regions, the threshold will become less than zero ($\theta < 0$) in $OT$ cases, ensuring that the synthetic instance will take the reference class (majority), i.e. $y_{cj} \leftarrow y_{rj}$
\end{itemize}

\begin{algorithm}[ht]
 \SetKwData{Left}{left}\SetKwData{This}{this}\SetKwData{Up}{up}
 \SetKwFunction{Random}{Random}
 \SetKwInOut{Input}{input}\SetKwInOut{Output}{output}
 \Input{seed instance: $(\bm{x}_s,\bm{y}_s)$, types of seed instance: $T_s$, reference instance: $(\bm{x}_r,\bm{y}_r)$, types of reference instance: $T_r$}
 \Output{synthetic instance: $(\bm{x}_c,\bm{y}_c)$}
\For{$j\leftarrow 1$ \KwTo $d$}
{
    $x_{cj} \leftarrow  x_{sj} +\Random(0,1)*(x_{rj}-x_{sj})$ \tcc*[r]{Random(0,1) return a random value $\in[0,1]$}
}
$d_s \leftarrow distance(\bm{x}_c,\bm{x}_s)$, $d_r \leftarrow distance(\bm{x}_c,\bm{x}_r)$ \;
$cd \leftarrow d_s/(d_s+d_r)$ \;
\For {$j\leftarrow 1$ \KwTo $q$}{
    \eIf{ $y_{sj}=y_{rj}$ }{
    $y_{cj} \leftarrow y_{sj}$ \;
    }{
        \If(\tcc*[h]{ensure $y_{sj}$ being minority class}){$T_{sj} = MJ$}{
            $s \longleftrightarrow r$ \tcc*[r]{swap indices of seed and reference instance}
            $cd \leftarrow 1-cd$ \;
        }
        \Switch{$T_{sj}$}{
            \lCase {$SF$}{
                $\theta \leftarrow 0.5$ ;  break
            }
            \lCase {$BD$}{
                $\theta \leftarrow 0.75$ ;  break
            }\lCase {$RR$}{
                $\theta \leftarrow 1+1e-5$ ;  break
            }\lCase {$OT$}{
                $\theta \leftarrow 0-1e-5$ ;  break
            }
        }
        \eIf{$cd \leqslant \theta$}{
            $y_{cj} \leftarrow y_{sj}$ \;
        }{
            $y_{cj} \leftarrow y_{rj}$ \;
        }
    }

}
\KwRet{$(\bm{x}_t,\bm{y}_t)$} \;
 \caption{CreateIns 
 }
  \label{al:GenerateInstance}
\end{algorithm}

\begin{figure*}
\begin{center}
\includegraphics[width=0.7\textwidth]{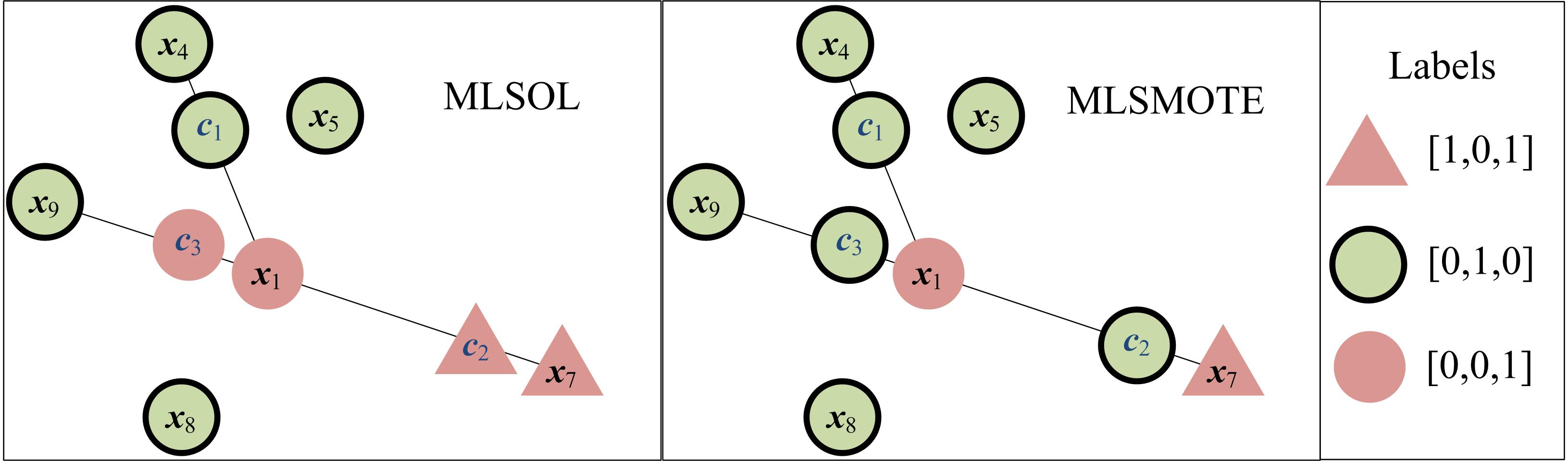}
\caption{The subset of dataset (b) concerning $\bm{x}_1$ and its $k$NNs as an example of MLSOL excelling MLSMOTE for label assignment for synthetic instances. $\bm{x}_1$ is the seed instance, $\bm{x_4}-\bm{x_9}$ are candidate reference instances ($kNN(\bm{x}_1)$), and $\bm{c}_*$ are possible synthetic examples.
}
\label{fig:MLSOLLabelAssignment}
\end{center}
\end{figure*}

Compared with MLROS and MLSMOTE, MLSOL performs a more comprehensive analysis by emphasizing on more difficult to learn instances and generating more diverse and well-labeled synthetic instances.
For dataset (b) in Fig.\ref{fig:ExampleDif}, MLROS would randomly replicate the red data points containing minority label ($l_3$) with equal chance. Likewise, the probabilities of each red data point to be selected as seed instance by MLSMOTE are equal. On the other hand, MLSOL is more likely to choose $\bm{x_1}$ as seed instance, because it is surrounded by more opposite class neighbours for $l_3$. With respect to the synthetic generation process, as shown in Fig.\ref{fig:MLSOLLabelAssignment}, MLSMOTE assigns label vector [0,1,0] to all synthetic instances, as decided by their neighbors. Conversely, MLSOL generates more diverse instances via assigning them labels according to their location. Furthermore, the synthetic instances $\bm{c}_2$ and $\bm{c}_3$ generated by MLSMOTE introduce noise, while MLSOL copies the labels of the nearest instance to the new examples. In conclusion, MLSMOTE generates new instances biased to the dominant class in the local area. In contrast, MLSOL is characterized by an efficient exploration and exploitation of the feature and labels space.

\subsection{Undersampling with MLUL
}
We propose a new Multi-Label UndersampLing method (MLUL) that makes minority class examples to be learned more easily via the removal of harmful examples. In MLUL, instead of picking up harmful examples to be deleted directly, we choose a subset of important examples and eliminate the rest.

In traditional undersampling approaches, majority class instances surrounding minority class examples are typically considered candidates for removal. However, this simple strategy becomes invalid in the case of multi-label data, where the same training example could be important to some of the labels but damaging for other labels. To deal with this issue, we need to consider two factors to evaluate the importance of an instance:
\begin{enumerate}
    \item Its local imbalance level.
    \item The influence of the instance $\bm{x}_i$ on its reverse nearest neighbours, R$k$NN. The R$k$NN of $\bm{x}_i$ is a group of instances where $\bm{x}_i$ belongs to their neighborhoods~\cite{TAO2004}, i.e.:
    \begin{equation}
    RkNN(\bm{x}_i) = \{\bm{x}_m| \bm{x}_i \in {\cal N}^k_m ), 1\leqslant m \leqslant n \}
    \label{eq:RkNN}
    \end{equation}
\end{enumerate}
The local imbalance of an instance could be measured by $\bm{w}$, as defined in Eq.\eqref{eq:wi2}. With respect to the second factor, we follow the principle that an instance is detrimental (beneficial) to an instance belonging to its R$k$NN if they have opposite (same) class for a label. To this direction, we introduce the influence quantity $u_i$ that measures the impact of instance $\bm{x}_i$ to its $RkNN(\bm{x}_i)$: 
\begin{equation}
u_i=\sum\limits_{j=1}^q \dfrac{\sum\limits_{\bm{x_m} \in RkNN(\bm{x_i})} (-1)^{\llbracket y_{ij} \neq y_{mj} \rrbracket} S_{mj} \llbracket S_{mj} \neq -1\rrbracket }{|RkNN(\bm{x_i})|}
\label{eq:ui}
\end{equation}
Influence is calculated by combining the influence degree of informative labels for $\bm{x}_i$. This is obtained by taking the algebraic sum of local imbalances $S_{mj}$ in all members of group $RkNN(\bm{x}_i)$, where $S_{mj}$ is added to or subtract from the accumulative influence, depending on whether $\bm{x}_i$ and $\bm{x}_m$ have equal or different class for label $l_j$. Obviously, influence can take positive ($u_i>0$) or negative ($u_i<0$) values, denoting that the impact of $\bm{x}_i$ on its $RkNN$ is useful or harmful, respectively. The larger the $|u_i|$, the more beneficial or harmful $\bm{x}_i$ is considered for its $RkNN$.
Specifically, we consider the importance of an instance as:
\begin{equation}
v_i=w_i+u_i-\min\{w_m+u_m\}_{1\leqslant m \leqslant n}
\label{eq:vi}
\end{equation}
where the subtraction of the third term ensures that all values in $\bm{v}\in \mathbb{R}^n$ are non-negative.

The pseudo-code of MLUL is shown in Algorithm \ref{al:MLUL}. Firstly, the number of retained instances is computed based on the undersampling ratio $p$ (line 1 in Algorithm \ref{al:MLUL}). Then, several auxiliary variables and the importance of instances $\bm{v}$ are calculated (lines 2-7 in Algorithm \ref{al:MLUL}). Subsequently, $RetNum$ instances are sampled without replacements, with the probability of selection being proportional to the importance it is associated with (Algorithm \ref{al:MLUL}, lines 8-13). Finally, the sampled instance subset $D'$ is retained and examples excluded from $D'$ are discarded.
In addition, we exemplify the advantage of MLUL over MLRUS with dataset (b) in Fig.\ref{fig:ExampleDif}. MLRUS would delete triangle points randomly, while MLUL would remove $\bm{x}_2$ and $\bm{x}_3$ with more probability than other triangle points because $\bm{x}_2$ and $\bm{x}_3$ are easier to be learned and hinder their R$k$NNs, i.e. the green triangle points with borderline for $l_2$ and red triangle points for $l_3$.

\begin{algorithm}
 \SetKwData{Left}{left}\SetKwData{This}{this}\SetKwData{Up}{up}
 \SetKwFunction{InitTypes}{InitTypes}\SetKwFunction{GenerateInstance}{GenerateInstance}
 \SetKwInOut{Input}{input}\SetKwInOut{Output}{output}

 \Input{multi-label data set: $D$, sampling ratio: $p$}
 \Output{new data set $D'$}
 $ RetNum \leftarrow |D|*(1-p)$ \tcc*[r]{number of instances to retain}
 $D' \leftarrow \emptyset$ \;
 Find the $k$NN and R$k$NN of each instance \;
 Calculate $\bm{C}$ according to Eq.\eqref{eq:Cij} \;
 Compute $\bm{S}$ according to Eq.\eqref{eq:Sij} \;
 Compute $\bm{w}$ according to Eq.\eqref{eq:wi2} \;
 Compute $\bm{u}$ according to Eq.\eqref{eq:ui}\;
 Compute $\bm{v}$ according to Eq.\eqref{eq:vi}\;
\While{$RetNum>0$}{
    Choose an instance $(\bm{x},\bm{y})$ from $D$ based on $\bm{v}$\;
    $D' \leftarrow D' \cap (\bm{x},\bm{y})$ \;
    $D  \leftarrow D  \setminus(\bm{x},\bm{y})$ \;
    $RetNum \leftarrow RetNum-1$ \;
}
\KwRet{$D'$} \;
 	
 \caption{MLUL}
  \label{al:MLUL}
\end{algorithm}

\subsection{Ensemble of Multi-Label Sampling (EMLS)}

Ensemble methods constitue an effective strategy to increase the overall accuracy and overcome over-fitting problems, but have not been leveraged in multi-label sampling approaches. To improve the robustness of the proposed multi-label sampling methods, we develop an Ensemble framework for Multi-Label Sampling (EMLS), where any multi-label sampling approach and classifier could be embedded. In EMLS, $M$ multi-label learning models are independently trained, where each model is built upon a sampled dataset generated by a multi-label sampling method with a different random seed. There are many random operations in existing and proposed multi-label learning sampling methods~\cite{Charte2015,Charte2015a}, which guarantee the diversity of the training set of each model in the ensemble framework by employing a different random seed. In addition, when our proposed sampling methods are used in the ensemble framework, setting different sampling ratio $p$ or number of neighbours $k$ offers another way to diversify the sampled datasets. Then the bipartition threshold of each label is decided by maximizing F-measure on the training set, as COCOA~\cite{Zhang2015a} and ECCRU3~\cite{Liu2018MakingImbalance} do.
Given a test example, the predicted relevance scores are calculated as the average of the relevance scores obtained from the $M$ models, and the labels whose relevance score is larger than the corresponding bipartition threshold are predicted as "1", and "0" otherwise.

\subsection{Complexity Analysis}
The complexity of searching $k$NN and R$k$NN of input instances is $O(n^2d+n^2k)$.
The complexity of computing auxiliary variables, such as $\bm{w}$, $\bm{T}$ and $\bm{u}$ is $O(nkq)$.
The complexity of sampling retained instances in MLUL is $O\left(pn\right)$.
Therefore, the overall complexity of MLUL is $O(n^2d+n^2k+nkq+pn)$.
The complexity of creating synthetic instances is $O(pn(q+d))$. Therefore, the overall complexity of MLSOL is $O(n^2d+n^2k+nkq+pn(q+d))$.
$k$NN searching is the most time-consuming part for both MLUL and MLSOL.
Compared with MLUL, MLSOL is more time-consuming due to the process of creating synthetic instances.

Let's define $\Theta_{s}(n,d,q,p)$ the complexity of a multi-label sampling approach, and $\Theta_{t}(n',d,q)$ and $\Theta_{p}(d,q)$ the complexity of training and prediction of multi-label learning method respectively where $n'$ is the size of the output sampled dataset. The complexity of EMLS is $O\left(M \left( \Theta_{s}(n,d,q,p)+\Theta_{t}(n',d,q)+n'\Theta_{p}(d,q) \right) \right)$ for training and $O(M\Theta_{p}(d,q))$ for prediction.
Generally, EMLS combined  with undersampling methods is much more efficient than with oversampling approaches, because undersampling methods are usually faster and output less number of instances than oversampling methods.

\section{Empirical Analysis \label{sec:experiments} }
In this section, we first describe the basic setup of experiments.
Then, experimental results are presented to show the effectiveness of $LImb$ and proposed approaches.
Lastly, the influence of parameters on our methods is analysed.

\subsection{Setup}

Table \ref{ta:Dataset} shows the 13 benchmark multi-label datasets used in our experimental study along with their global and local imbalance levels. All datasets are available online at Mulan\footnote{\href{http://mulan.sourceforge.net/datasets-mlc.html}{http://mulan.sourceforge.net/datasets-mlc.html}}.
In textual data sets with more than 1000 features, we applied a simple feature selection approach that retains the top 10$\%$ (bibtex, enron, medical) or top 1$\%$ (rcv1subset1, rcv1subset2, yahoo-Arts1, yahoo-Business1) of the features ordered by number of non-zero values (i.e. frequency of appearance). We remove labels containing only one minority class instance, because when splitting the dataset into training and test sets, there may be only majority class instances of those extremely imbalanced labels in the training set.

\begin{table*}[t]  
\centering
\caption{The 13 multi-label datasets used in this study. Columns $n$, $d$, $q$ denote the number of instances, features and labels respectively, $LC$ the label cardinality. The $k=5$ for $LImb$.
}
\label{ta:Dataset}
\begin{tabular}{@{}cccccccccccc@{}}
\toprule
Dataset & Domain & $n$ & $d$ & $q$ & $LC$ & $MeanIR$ & $CVIR$ & $MeanImR$ & $CVImR$ & $SCUMBLE$ & $LImb$ \\ \midrule
bibtex & text & 7395 & 183 & 159 & 2.402 & 12.5 & 0.4051 & 87.7 & 0.4097 & 0.0938 & 0.8816 \\
cal500 & music & 502 & 68 & 174 & 26 & 20.6 & 1.087 & 22.3 & 1.129 & 0.3372 & 0.8485 \\
corel5k & image & 5000 & 499 & 347 & 3.517 & 117 & 1.128 & 522 & 1.13 & 0.3917 & 0.9725 \\
enron & text & 1702 & 100 & 52 & 3.378 & 57.8 & 1.482 & 107 & 1.496 & 0.3024 & 0.844 \\
flags & image & 194 & 19 & 7 & 3.392 & 2.255 & 0.7648 & 2.753 & 0.7108 & 0.0606 & 0.5163 \\
genbase & biology & 662 & 1186 & 24 & 1.248 & 20.6 & 1.269 & 78.8 & 1.286 & 0.0266 & 0.2112 \\
medical & text & 978 & 144 & 35 & 1.245 & 39.1 & 1.107 & 143 & 1.115 & 0.0415 & 0.7438 \\
rcv1subset1 & text & 6000 & 472 & 101 & 2.88 & 54.5 & 2.081 & 236 & 2.089 & 0.2237 & 0.896 \\
rcv1subset2 & text & 6000 & 472 & 101 & 2.634 & 45.5 & 1.715 & 191 & 1.724 & 0.2092 & 0.887 \\
scene & image & 2407 & 294 & 6 & 1.074 & 1.254 & 0.1222 & 4.662 & 0.1485 & 0.0003 & 0.2633 \\
yahoo-Arts1 & text & 7484 & 231 & 25 & 1.654 & 25 & 2.444 & 101 & 2.468 & 0.0594 & 0.8523 \\
yahoo-Business1 & text & 11214 & 219 & 28 & 1.599 & 249 & 2.447 & 286 & 2.453 & 0.1252 & 0.8603 \\
yeast & biology & 2417 & 103 & 14 & 4.237 & 7.197 & 1.884 & 8.954 & 1.997 & 0.1044 & 0.5821 \\ \bottomrule
\end{tabular}
\end{table*}

Four multi-label sampling methods, namely MLRUS, MLROS~\cite{Charte2015}, MLSMOTE~\cite{Charte2015a} and RHwRSMT~\cite{Charte2019a}, are used for comparison, among which the first one is an undersampling method, while the other three are oversampling methods. The ensemble versions of the proposed MLUL and MLSOL methods, denoted as EMLUL and EMLSOL, are compared with the ensemble versions of competing approaches, namely EMLRUS, EMLROS, EMLSMOTE and ERHwRSMT respectively. Furthermore, the base learning algorithm without employing any sampling method, denoted as {\em Default}, is also used for comparison purposes.
In MLUL, MLSOL, MLSMOTE, RHwRSMT, the number of nearest neighbours $k$ is set to 5 and the Euclidean distance is used to measure the distance between the examples.
The sampling ratio $p$ is set to 0.3 for MLSOL and 0.1 for MLUL, MLRUS and MLROS. In RHwRSMT, the threshold for decoupling an instance is set to $SCUMBLE$.
The ensemble size $M$ is set to 5 for all ensemble methods.
In addition, six multi-label learning methods are employed as base learning methods, comprising four standard multi-label learning methods (BR~\cite{Boutell2004}, MLkNN~\cite{zhang2007ml}, CLR~\cite{Furnkranz2008}, RAkEL~\cite{tsoumakas2011random}), as well as two state-of-the-art methods addressing the class imbalance problem (COCOA~\cite{Zhang2015a} and ECCRU3~\cite{Liu2018MakingImbalance}).


Three widely used imbalance aware evaluation metrics are leveraged to measure the performance of methods:
\begin{itemize}
    \item Macro-averaged {\em F-measure},
    \item Macro-averaged {\em AUC-ROC} (area under the receiver operating characteristic curve), and
    \item Macro-averaged {\em AUCPR} (area under the precision-recall curve)
\end{itemize}
For simplicity, we omit the ``macro-averaged" in further references to these metrics within the rest of this paper.
To examine the statistical significance of the differences among the competing methods, the Friedman test, followed by the Wilcoxon signed rank test with Bergman-Hommel's correction at the 5\% level is employed~\cite{Garcia2008,Benavoli2016}.

The experiments were conducted on a machine with 4$\times$10-core CPUs running at 2.27 GHz. We apply $5 \times 2$-fold cross validation with multi-label stratification~\cite{Sechidis2011} to each dataset and the average results are reported. The implementation of our approach
is publicly available at Mulan's GitHub repository\footnote{\href{https://github.com/tsoumakas/mulan/tree/master/mulan}{https://github.com/tsoumakas/mulan/tree/master/mulan}}. The default parameters are used for base learners.

\subsection{Effectiveness of $LImb$}

More difficult dataset usually leads to lower performance of predicting method. If a measure is negatively correlated with the predicting performance, it is positively correlated with the difficult level of dataset.
Therefore, to investigate which imbalance measure can reveal the difficulty of multi-label dataset,
we calculate the Pearson correlation coefficients ($\rho$) between each measure and the performance of each base approach on the 13 datasets that are listed in Table \ref{ta:Dataset}. The results are presented in Table \ref{ta:CorMeasurementPerformance}.
At first, we notice that all measures are negatively correlated with the performance. 
In most cases, $LImb$ is the most significantly correlated measurement with $\rho$ round -0.9, followed by $SCUMBLE$ whose $\rho$ is nearly -0.7. $SCUMBLE$ is the most correlated measure in terms of AUC-ROC of CLR and COCOA, which both belong to the pairwise transformation strategy. Yet, the strength of these correlations are not significant.
Overall, $LImb$, which reflects the local label imbalance, rather than the global label imbalance based measures,
is the most effective measure to assess the difficulty of multi-label dataset.

\begin{table*}[t]  
\centering
\caption{The Pearson correlation coefficients ($\rho$) between measures and performances on 13 datasets. Parentheses denote the rank of the corresponding $\rho$ in each row. The significant $\rho$ not in the range of critical values [-0.684,0.684] with $\alpha=0.01$ are boldfaced.}
\label{ta:CorMeasurementPerformance}
\begin{tabular}{cccccccc}
\toprule
 Metric & Base & \textit{MeanIR} & \textit{CVIR} & \textit{MeanImR} & \textit{CVImR} & \textit{SCUMBLE} & \textit{LImb} \\ \midrule
\multirow{6}{*}{F-measure} & BR & -0.401(4) & -0.348(5) & -0.52(3) & -0.347(6) & \textbf{-0.692(2)} & \textbf{-0.959(1)} \\
 & MLkNN & -0.428(6) & -0.493(4) & -0.552(3) & -0.487(5) & -0.648(2) & \textbf{-0.964(1)} \\
 & CLR & -0.387(4) & -0.368(5) & -0.513(3) & -0.367(6) & \textbf{-0.696(2)} & \textbf{-0.962(1)} \\
 & RAkEL & -0.412(4) & -0.385(5) & -0.537(3) & -0.383(6) & \textbf{-0.686(2)} & \textbf{-0.971(1)} \\
 & COCOA & -0.433(4) & -0.377(5.5) & -0.56(3) & -0.377(5.5) & \textbf{-0.717(2)} & \textbf{-0.959(1)} \\
 & ECCRU3 & -0.428(6) & -0.493(4) & -0.552(3) & -0.487(5) & -0.648(2) & \textbf{-0.964(1)} \\ \midrule
\multirow{6}{*}{AUC-ROC} & BR & -0.314(6) & -0.334(3.5) & -0.318(5) & -0.334(3.5) & -0.638(2) & \textbf{-0.852(1)} \\
 & MLkNN & -0.318(6) & -0.412(3) & -0.359(5) & -0.404(4) & -0.674(2) & \textbf{-0.828(1)} \\
 & CLR & -0.14(5) & -0.169(4) & -0.062(6) & -0.174(3) & -0.532(1) & -0.353(2) \\
 & RAkEL & -0.337(6) & -0.348(4.5) & -0.41(3) & -0.348(4.5) & \textbf{-0.688(2)} & \textbf{-0.918(1)} \\
 & COCOA & -0.181(5) & -0.247(4) & -0.154(6) & -0.255(3) & -0.645(1) & -0.487(2) \\
 & ECCRU3 & -0.318(6) & -0.412(3) & -0.359(5) & -0.404(4) & -0.674(2) & \textbf{-0.828(1)} \\ \midrule
\multirow{6}{*}{AUCPR} & BR & -0.417(4) & -0.358(5) & -0.528(3) & -0.357(6) & -0.637(2) & \textbf{-0.945(1)} \\
 & MLkNN & -0.449(4) & -0.435(5) & -0.572(3) & -0.427(6) & -0.645(2) & \textbf{-0.992(1)} \\
 & CLR & -0.462(4) & -0.378(5) & -0.566(3) & -0.374(6) & \textbf{-0.693(2)} & \textbf{-0.967(1)} \\
 & RAkEL & -0.424(4) & -0.406(5) & -0.56(3) & -0.403(6) & -0.666(2) & \textbf{-0.977(1)} \\
 & COCOA & -0.461(4) & -0.413(5) & -0.587(3) & -0.41(6) & \textbf{-0.692(2)} & \textbf{-0.976(1)} \\
 & ECCRU3 & -0.449(4) & -0.435(5) & -0.572(3) & -0.427(6) & -0.645(2) & \textbf{-0.992(1)} \\ \bottomrule
\end{tabular}
\end{table*}

\begin{figure}
\begin{center}
\includegraphics[width=0.45\textwidth]{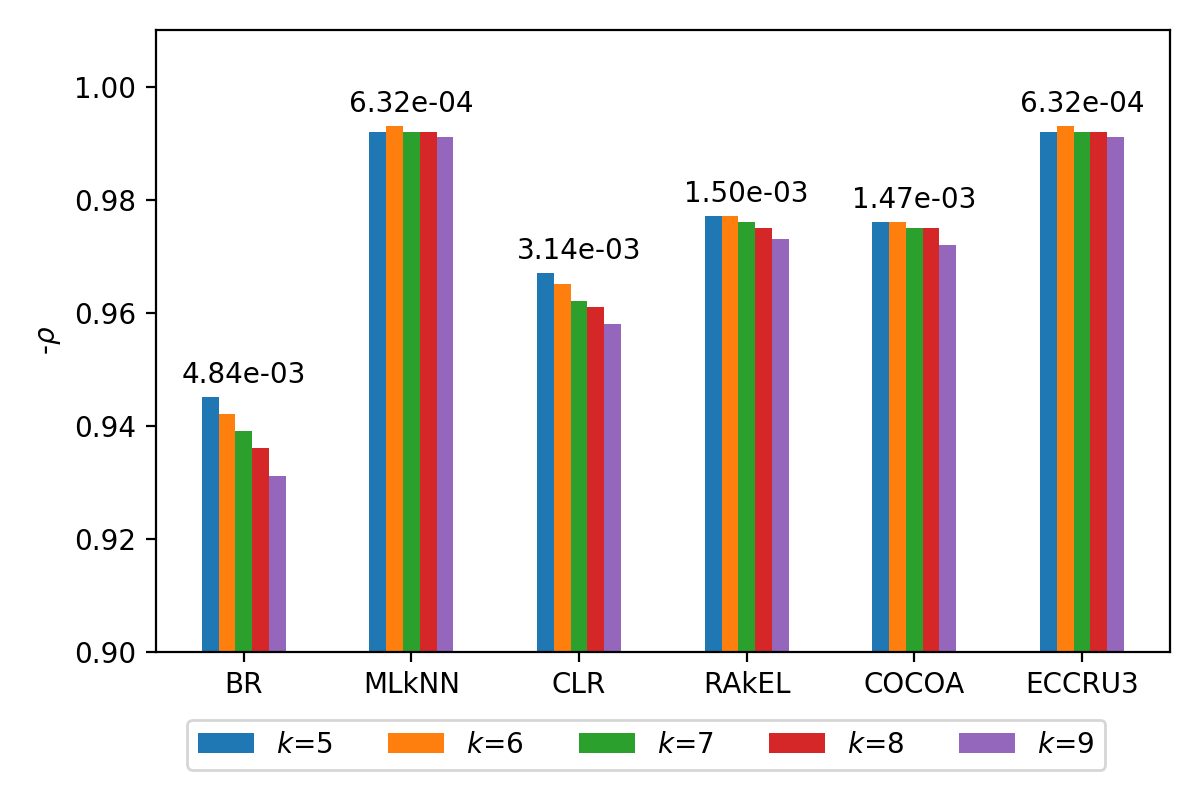}
\caption{Pearson correlation coefficients between $LImb$ with various $k$ and AUCPR on 13 dataset sets.}
\label{fig:Rho_Ks}
\end{center}
\end{figure}

Furthermore, we analyse the influence of the number of neighbours $k$ on the Pearson correlation coefficient between $LImb$ and model performance.
Fig \ref{fig:Rho_Ks} shows the negative $\rho$ in terms of AUCPR on the 13 dataset sets under various $k$, where the value on the top of each base learner is the standard variance of the corresponding 5 coefficients calculated with different $k$. Obviously, all $\rho$ values are less than -0.9, implicating the significance of the correlation. Besides, all standard variances are less than 0.005, indicating the insensitivity of $\rho$ with respect to the changing of $k$. Therefore the relation between $LImb$ and performance of multi-label learning approaches is effective and stable, regardless of the actual $k$.

\subsection{Results of Sampling Methods}

Table \ref{ta:Results_Sampling} shows the average rank of each method as well as its significant wins/losses versus each one of the rest of the methods for each of the three evaluation metrics and each of the six base multi-label methods. 

Three out of four oversampling approaches, namely MLROS, MLSMOTE and MLSOL, achieve the top 3 average ranks in most cases. Specifically, MLSOL is the best method in 7 cases, followed by MLROS and MLSMOTE which has the best average rank in 5 and 4 cases, respectively. Although MLSOL is more effective than other competing methods in more cases, there is no single oversampling method that achieves the best result for all base learners in terms of all metrics. MLSOL has the most total significant wins (18) in F-measure and doesn't suffer any significant loss in any metric. MLROS has the most total significant wins (12) in AUC-ROC, followed by MLSOL (11). In terms of AUCPR, the Default approach, along with MLROS, have the most total significant wins (9), without suffering any significant loss, indicating that multi-label sampling methods are not very effective in AUCPR metric which is considered as the most appropriate measure in the context of class imbalance.

With respect to undersampling approaches, the proposed method, MLUL, is better than MLRUS, because MLUL retains more important instances via considering local imbalance. Nevertheless, both of the two undersampling methods are inferior to the Default approach, which is due to the inevitable information loss caused by removing instances.
RHwRSMT is the worst because of the additional bewilderment yielded by REMEDIAL, i.e. there are several pairs of instances with the same features and disparate labels.

It should be also noticed that ECCRU3 and COCOA achieve the best average rank and most significant wins in terms of F-measure, which is mainly due to the utilization of the bipartition threshold selection strategy that maximizes F-measure for each label on the training set.
Furthermore, none of the sampling methods are able to significantly improve the performance of ECCRU3 in any measure. This is expected to a certain degree, as ECCRU3 is an imbalance aware method.

Overall, oversampling methods could improve the performance in several cases but are not very effective in terms of AUCPR and for ECCRU3, which directly tackles the class imbalance issue. Undersampling methods are worse than base learners as well as oversampling methods. The proposed MLSOL is slightly better than other state-of-the-art oversampling approaches in several cases and the proposed MLUL is better than the existing udnersampling method.

\begin{table*}[t]  
\centering
\caption{Average rank of the compared sampling methods using 6 base learners in terms of three evaluation metrics. The parenthesis ($n_1$/$n_2$) indicates the corresponding method is significantly superior to $n_1$ methods and inferior to $n_2$ methods based on the Wilcoxon signed rank test with Bergman-Hommel's correction at the 5\% level. The best methods are highlighted by boldface.
}
\label{ta:Results_Sampling}
\begin{tabular}{@{}ccccccccc@{}}
\toprule
Metric & Base & Default & MLRUS & MLUL & MLROS & MLSMOTE & RHwRSMT & MLSOL \\ \midrule
\multirow{7}{*}{F-measure} & BR & 4.12(1/3) & 5.23(1/3) & 4.69(1/3) & 2.85(4/0) & \textbf{1.92(4/0)} & 7(0/6) & 2.19(4/0) \\
 & MLkNN & 4.15(2/3) & 5.46(1/5) & 4.54(2/3) & 3(4/2) & 2.54(5/1) & 6.92(0/6) & \textbf{1.38(6/0)} \\
 & CLR & 3.31(2/1) & 5.31(1/4) & 4.58(1/2) & 3.42(2/0) & 2.54(3/0) & 6.92(0/6) & \textbf{1.92(4/0)} \\
 & RAkEL & 4.04(2/3) & 5.46(1/5) & 4.77(2/3) & 2.77(4/0) & 2.27(4/0) & 7(0/6) & \textbf{1.69(4/0)} \\
 & COCOA & \textbf{2.5(2/0)} & 5.38(0/2) & 3.23(2/0) & 3.35(1/0) & 3.77(1/0) & 5.88(0/4) & 3.88(0/0) \\
 & ECCRU3 & \textbf{2(2/0)} & 4.73(0/1) & 3.69(0/0) & 3.54(1/0) & 3.46(1/0) & 5.85(0/3) & 4.73(0/0) \\ \cmidrule(l){2-9}
 & \textit{Ave(Total)} & 3.35(11/10) & 5.26(4/20) & 4.25(8/11) & 3.16(16/2) & 2.75(18/1) & 6.6(0/31) & \textbf{2.63(18/0)} \\ \midrule
\multirow{7}{*}{AUC-ROC} & BR & 4.23(0/1) & 5.31(0/3) & 4.62(0/1) & 3.46(2/0) & 2.92(2/0) & 5.54(0/3) & \textbf{1.92(4/0)} \\
 & MLkNN & 3.08(0/0) & 5.73(0/0) & 4.04(0/0) & 3.5(0/0) & \textbf{2.46(1/0)} & 5.46(0/1) & 3.73(0/0) \\
 & CLR & 3(2/0) & 5.85(0/2) & 5.08(0/1) & \textbf{2.15(1/0)} & 3.58(1/0) & 4.54(0/1) & 3.81(0/0) \\
 & RAkEL & 3.77(2/1) & 5.31(1/5) & 4.38(2/2) & 2.96(2/0) & 2.81(3/0) & 7(0/6) & \textbf{1.77(4/0)} \\
 & COCOA & 3.85(2/2) & 5.96(0/3) & 5.5(0/3) & \textbf{2.04(4/0)} & 3.62(0/0) & 4.15(0/1) & 2.88(3/0) \\
 & ECCRU3 & 3.31(2/0) & 5.65(0/2) & 5.15(0/2) & \textbf{2.35(3/0)} & 2.77(0/0) & 4.77(0/1) & 4(0/0) \\  \cmidrule(l){2-9}
 & \textit{Ave(Total)} & 3.54(8/4) & 5.64(1/15) & 4.8(2/9) & \textbf{2.74(12/0)} & 3.03(7/0) & 5.24(0/13) & 3.02(11/0) \\ \midrule
\multirow{7}{*}{AUCPR} & BR & 3.69(1/0) & 4.85(1/0) & 4.77(0/0) & 3.31(1/0) & \textbf{2.58(1/0)} & 6.12(0/5) & 2.69(1/0) \\
 & MLkNN & 3.54(0/0) & 5.38(0/0) & 4.5(0/0) & 4.04(0/0) & 3\textbf{(1/0)} & 5.04(0/1) & \textbf{2.5}(0/0) \\
 & CLR & 2.69(2/0) & 5.69(0/0) & 5.08(0/2) & \textbf{2(2/0)} & 3.81(0/0) & 4.81(0/2) & 3.92(0/0) \\
 & RAkEL & 3.12(3/0) & 5.31(1/3) & 4.62(1/3) & 3(1/0) & 2.65(3/0) & 7(0/6) & \textbf{2.31(3/0)} \\
 & COCOA & 3.85(1/0) & 5.77(0/2) & 5.31(0/1) & \textbf{2.15(3/0)} & 3.65(0/0) & 4.23(0/1) & 3.04(0/0) \\
 & ECCRU3 & 3.23(2/0) & 6(0/3) & 5.62(0/3) & 2.81(2/0) & \textbf{2.77(2/0)} & 4(0/0) & 3.58(0/0) \\  \cmidrule(l){2-9}
 & \textit{Ave(Total)} & 3.35(9/0) & 5.5(2/8) & 4.98(1/9) & \textbf{2.89(9/0)} & 3.08(7/0) & 5.2(0/15) & 3.01(4/0) \\ \bottomrule
\end{tabular}
\end{table*}

\subsection{Results of Ensemble Methods}

In this part, we examine the effectiveness of EMLS. We compare EMLS coupled with the six sampling approaches, as well as the Default approach. Average ranks and statistical test results are shown in Table \ref{ta:Results_Ensemble}.

Firstly, we observe that embedding a sampling method in the EMLS approach can significantly improve the performance of using the sampling method alone for all base learners and in all evaluation metrics. This verifies the known effectiveness of resampling approaches in reducing the error, in particular via reducing the variance component of the expected error~\cite{Hastie2016}.
EMLUL is the best method in terms of F-measure, and EMLSOL is the top one in terms of AUC-ROC and AUCPR. Both EMLSOL and EMLUL are not significantly inferior to the other five comparing methods. The single significant loss of EMLSOL is caused by EMLUL and the 5 significant losses of EMLUL are attributed to EMLSOL.
EMLROS and EMLRUS are usually the third and fourth methods following our two approaches, and EMLSMOTE comes next. The top five ensemble approaches are better than Default, with two exceptions: EMLSMOTE is worse than the Default approach for COCOA and ECCRU3 in terms of F-measure. ERHwRSMT is the worst ensemble method, even worse than Default in some cases (i.e. for RAkEL, COCOA and ECCRU3 in F-measure). Furthermore, none of the ensemble methods is significantly better than COCOA and ECCRU3 in terms of F-measure, which is also due to the employment of the bipartition threshold optimization strategy on F-measure.


Diversity is an important factor of the effectiveness of ensemble approaches~\cite{Zhou2012}. We therefore analyze the diversity of ensembling the different multi-label sampling methods with EMLS to shed light on the reasons for the superior performance achieved by our proposed methods. To assess the ensemble diversity, a widely used pairwise diversity measure, called \textit{disagreement}~\cite{Zhou2012}, is employed. Although the original definition of disagreement aims to evaluate diversity among binary classifiers, it could be easily applied to the multi-label learning scenario via a simple transformation. Given a test set including $n_t$ instances and two prediction matrices $Y^i\in \{0,1\}^{n_t \times q}$ and $Y^j\in \{0,1\}^{n_t \times q}$ obtained from two multi-label learning methods $f^i$ and $f^j$ respectively, the disagreement of the two learners is calculated as the proportion of different predictions in the two matrices:
\begin{equation}
disa_{ij}=\frac{\sum_{m=1}^{n_t}\sum_{l=1}^{q}\llbracket Y^{i}_{ml} \neq Y^{j}_{ml} \rrbracket}{n_tq}
\end{equation}
The disagreement of an ensemble model is computed as the average disagreement of every pair of multi-label base learners embedded in the ensemble framework:
\begin{equation}
disa=\frac{2\sum_{i=1}^{M-1}\sum_{j=i+1}^{M}disa_{ij}}{M(M-1)}
\end{equation}
The $disa$ measure takes values in $[0,1]$. The larger the $disa$ mesaure, the more diverse the ensemble model.

Table \ref{ta:Results_Diversity} lists the average rank of disagreement of ensemble models with six resampling approaches on the 13 datasets.
In Table \ref{ta:Results_Diversity} we seem that EMLSOL has the lowest $disa$ value for standard base methods, while it is the third best method for imbalance aware base methods (COCOA and ECCRU3), ahead of the other three oversampling methods. Although MLSOL does not have many significant advantages over MLROS and MLSMOTE, EMLSOL outperforms EMLROS and EMLSMOTE, because MLSOL generates more diverse synthetic instances. EMLUL and EMLRUS are more diverse than EMLROS and EMLSMOTE, which results in that EMLUL (EMLRUS) is better than (comparable with) EMLROS and EMLSOMTE, despite the inferior performance of the single undersampling method. The outputs of each model in ERHwRSMT are most consistent, which also results in its worse performance.  Overall, our proposed sampling methods embedded within EMLS excel other competing approaches mainly because MLUL and MLSOL can benefit more from the ensemble framework via outputting more diverse sampled datasets.


\begin{table*}[t]  
\centering
\caption{Average rank of the compared ensemble of resampling methods using 6 base learners in terms of three evaluation metrics. The parenthesis ($n_1$/$n_2$) indicates the corresponding method is significantly superior to $n_1$ methods and inferior to $n_2$ methods based on the Wilcoxon signed rank test with Bergman-Hommel's correction at the 5\% level. The "*" following the average rank denotes that the ensemble methods significantly outperform their corresponding signal sampling approaches.
The best methods are highlighted by boldface.
}
\label{ta:Results_Ensemble}
\begin{tabular}{@{}ccccccccc@{}}
\toprule
Metric & Base & Default & EMLRUS & EMLUL & EMLROS & EMLSMOTE & ERHwRSMT & EMLSOL \\ \midrule
\multirow{7}{*}{F-measure} & BR & 6.38(0/5) & 4.23*(2/1) & 3.46*(2/1) & 3.69*(1/0) & 3.08*(2/1) & 5.62*(0/4) & \textbf{1.54*(5/0)} \\
 & MLkNN & 7(0/6) & 3.38*(1/0) & 3*(1/0) & 2.85\textbf{*(2/0)} & \textbf{2.69*(2/0)} & 5.23*(1/2) & 3.85*(1/0) \\
 & CLR & 6.15(0/6) & 3.15*(1/0) & \textbf{2.81*(3/0)} & 3.69*(1/0) & 4.5*(1/1) & 4.31*(1/1) & 3.38*(1/0) \\
 & RAkEL & 6.15(0/5) & 3.85\textbf{*(2/0)} & 3.38\textbf{*(2/0)} & 3.08\textbf{*(2/0)} & 3.46\textbf{*(2/0)} & 6.31*(0/5) & \textbf{1.77*(2/0)} \\
 & COCOA & 5.15(0/0) & 2.92*(2/0) & \textbf{2.23*(3/0)} & 3.62*(2/0) & 5.23*(0/4) & 5.38*(0/3) & 3.46*(1/1) \\
 & ECCRU3 & 5(0/0) & 2.69*(1/0) & \textbf{2.31*(2/0)} & 3.38*(1/0) & 5.62*(0/2) & 5.46*(0/3) & 3.54*(1/0) \\ \cmidrule(l){2-9}
 & \textit{Ave(Total)} & 5.97(0/22) & 3.37(9/1) & \textbf{2.87(13/1)} & 3.39(9/0) & 4.1(7/8) & 5.39(2/18) & 2.92(11/1) \\ \midrule
\multirow{7}{*}{AUC-ROC} & BR & 6.23(0/5) & 3.77*(2/2) & 3*(3/1) & 3.69*(2/1) & 4.15*(2/1) & 6.08*(0/5) & \textbf{1.08*(6/0)} \\
 & MLkNN & 5.62(0/3) & 3.31*(2/0) & 3.15\textbf{*(3/0)} & 5.38*(0/3) & 3.23*(1/0) & 5.54*(0/3) & \textbf{1.77*(3/0)} \\
 & CLR & 5.92(0/3) & 4*(0/2) & \textbf{2.35*(4/0)} & 3.12*(2/0) & 4.65*(1/2) & 5.5*(0/4) & 2.46\textbf{*(4/0)} \\
 & RAkEL & 5.77(1/5) & 3.81*(2/2) & 2.5*(4/1) & 3.81*(2/1) & 3.96*(2/2) & 6.88*(0/6) & \textbf{1.27*(6/0)} \\
 & COCOA & 6.46(0/5) & 3.69*(2/1) & 2.38\textbf{*(4/0)} & 2.38*(3/0) & 5.23*(1/3) & 5.54*(0/4) & \textbf{2.31}*(3/0) \\
 & ECCRU3 & 6.31(0/4) & 3.42*(3/0) & 2.69\textbf{*(3/0)} & 2.38*(3/0) & 5.08*(1/4) & 6.15*(0/5) & \textbf{1.96*(3/0)} \\ \cmidrule(l){2-9}
 & \textit{Ave(Total)} & 6.05(1/25) & 3.67(11/7) & 2.68(21/2) & 3.46(12/5) & 4.38(8/12) & 5.95(0/27) & \textbf{1.81*(25/0)} \\ \midrule
\multirow{7}{*}{AUCPR} & BR & 6.12(0/5) & 3.73*(2/2) & 2.96*(4/1) & 3.69*(2/1) & 3.96*(2/2) & 6.38*(0/5) & \textbf{1.15*(6/0)} \\
 & MLkNN & 6.15(0/5) & 2.85*(2/0) & 3.27*(2/0) & 4.88*(1/1) & 3.88*(2/1) & 5.58*(0/4) & \textbf{1.38*(4/0)} \\
 & CLR & 6.19(0/4) & 3.38*(2/0) & 2.54\textbf{*(3/0)} & 2.62*(2/0) & 5.12*(0/2) & 6*(0/4) & \textbf{2.15*(3/0)} \\
 & RAkEL & 6.04(0/5) & 3.5*(2/1) & 2.5*(3/1) & 3.69*(2/0) & 4.42*(2/2) & 6.54*(0/5) & \textbf{1.31*(5/0)} \\
 & COCOA & 6.54(0/5) & 3.54*(3/2) & 2.19\textbf{*(4/0)} & 2.73*(3/0) & 5.31*(1/4) & 5.69*(0/4) & \textbf{2*(4/0)} \\
 & ECCRU3 & 6.38(0/4) & 3.27\textbf{*(3/0)} & 2.38\textbf{*(3/0)} & 2.65\textbf{*(3/0)} & 5.15*(1/4) & 6.08*(0/5) & \textbf{2.08*(3/0)} \\ \cmidrule(l){2-9}
 & \textit{Ave(Total)} & 6.24(0/28) & 3.38(14/5) & 2.64(19/2) & 3.38(13/2) & 4.64(8/15) & 6.05(0/27) & \textbf{1.68(25/0)} \\ \bottomrule
\end{tabular}
\end{table*}

\begin{table}
\centering
\scriptsize
\caption{Average rank of disagreement ($disa$) of multi-label sampling ensemble approaches on 13 datasets.
}
\label{ta:Results_Diversity}
\setlength\tabcolsep{2pt} 
\begin{tabular}{@{}ccccccc@{}}
\toprule
Base & EMLRUS & EMLUL & EMLROS & EMLSMOTE & ERHwRSMT & EMLSOL \\ \midrule
BR & 3.08 & 2.77 & 4.15 & 4.23 & 5.62 & \textbf{1.15} \\
MLkNN & 2.96 & 2.81 & 4.23 & 4.42 & 5.35 & \textbf{1.23} \\
CLR & 2.88 & 2.65 & 4.31 & 4.31 & 5.69 & \textbf{1.15} \\
RAkEL & 2.81 & 2.58 & 4.38 & 4.38 & 5.77 & \textbf{1.08} \\
COCOA & \textbf{1.77} & 2 & 3.77 & 5.12 & 5.58 & 2.77 \\
ECCRU3 & 2 & \textbf{1.46} & 3.77 & 5.42 & 5.19 & 3.15 \\ \midrule
\textit{Ave} & 2.24 & 2.67 & 3.54 & 4.46 & 4.77 & \textbf{2.21} \\
\bottomrule
\end{tabular}
\end{table}

\subsection{Parameter Analysis}
\begin{figure*}[!t]
\centering
\subfloat[enron]{\includegraphics[width=0.4\textwidth]{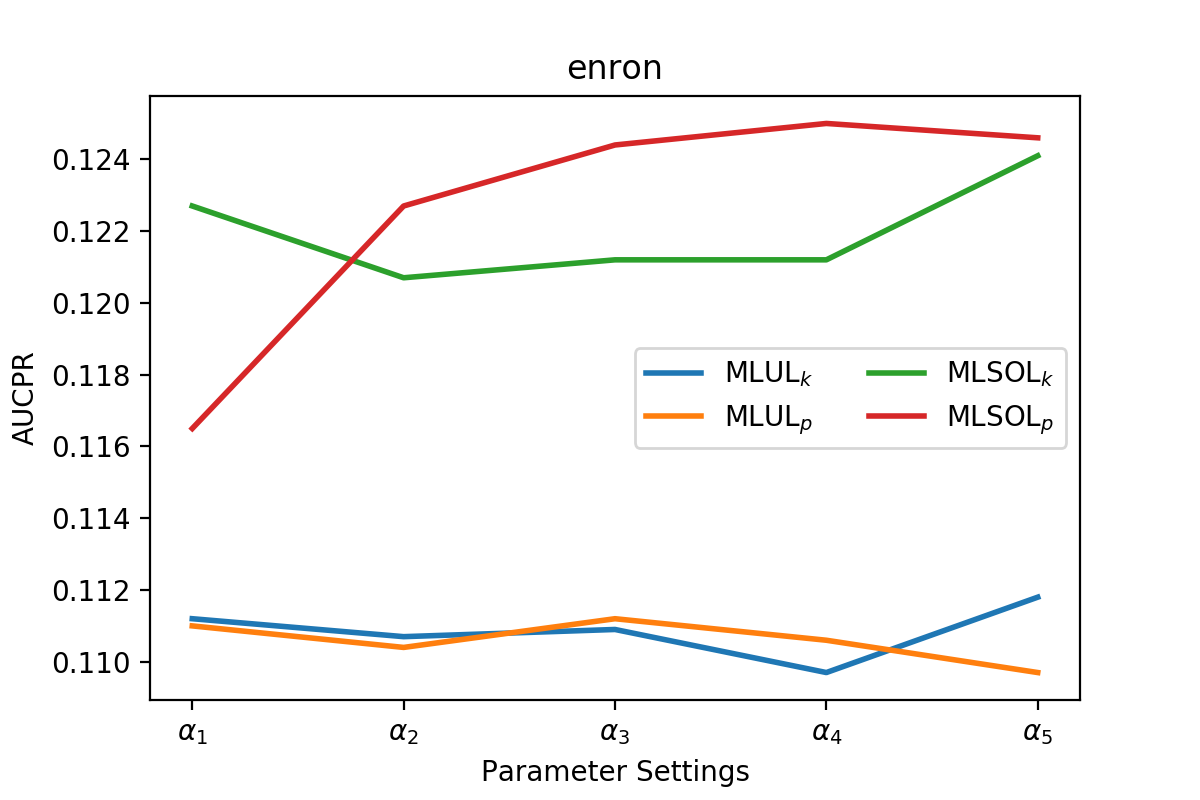}
\label{fig:enron}}
\subfloat[scene]{\includegraphics[width=0.4\textwidth]{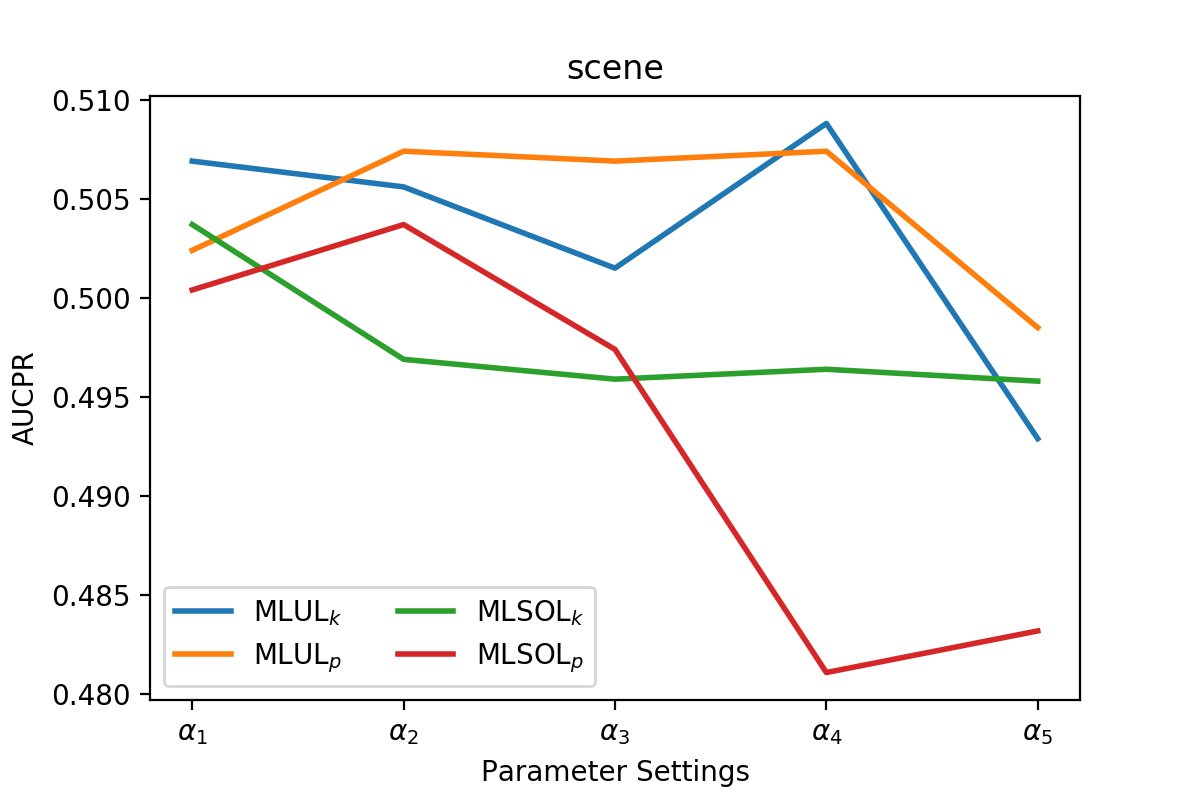}
\label{fig:scene}}
\caption{Results of MLSOL and MLUL with various parameter settings on enron and scene datasets in terms of AUCPR}
\label{fig:VarParameters}
\end{figure*}

An additional study has been made in order to investigate the influence of parameters, namely the number of neighbours $k$ and sampling ratio $p$, on the proposed methods, MLSOL and MLUL. In the experiments, we use different settings for one parameter, while keeping others unchanged at the setting illustrated in section 4.1. We vary $k=\{5,6,7,8,9\}$ for both sampling approaches, $p=\{0.01,0.05,0.1,0.15,0.2\}$ for MLUL and $p=\{0.1,0.3,0.5,0.7,0.9\}$ for MLSOL. The results on the enron dataset with $LImb$=0.844 and the scene dataset with $LImb$=0.2633 using BR as base learner in terms of AUCPR are shown in Fig.\ref{fig:VarParameters}.

Generally, MLSOL is more effective in the more difficult dataset (enron), while MLUL outperforms MLSOL in the scene dataset that could be learned more easily. Besides, compared with MLSOL, MLUL is more insensitive to the parameter settings.

We start by we focusing our study on parameter $p$. In the enron dataset, with the increase of $p$, the performance of MLSOL improves, which indicates that generating more synthetic instances contributes to lessening the hardness of the difficult dataset. In the scene dataset, MLSOL achieves the best result when $p=0.3$, but it fails when excess instances are added.
MLUL can slightly improve the performance by removing small parts of harmful instances for both datasets. However, when too many instances are removed, it's inevitable to lose informative instances, which results in performance decrease.
With respect to $k$, although the optimal settings of MLSOL and MLUL on the two datasets are different, $k=5$ seems a relative good setting for both approaches on all datasets.
However, discovering the optimum value of $k$ and designing neighboring regions in an optimal way, constitute a possible direction in our future research.

Then, we examine two alternative ways, setting various $p$ and $k$ for MLUL and MLSOL, to increase the diversity of EMLS. Specifically, E$_k$MLSOL (E$_k$MLUL) contains five MLSOL (MLUL) models and the $k$ of each one is set as $k=\{5,6,7,8,9\}$ respectively. Analogously, E$_p$MLSOL (E$_p$MLUL) denotes the embedded MLSOLs (MLULs) with $p=\{0.1,0.3,0.5,0.7,0.9\}$ ($p=\{0.01,0.05,0.1,0.15,0.2\}$), respectively. The statistical test results are shown in Table \ref{ta:Results_EMLSpk}. The E$_p$MLSOL and E$_k$MLSOL manage to enhance the diversity of the EMLS, therefore they significantly outperform the EMLSOL. Besides, the higher $p$ promotes the performance of MLSOL for difficult datasets, which also contributes to the improvement of E$_p$MLSOL. In contrast, there are a few significant differences between the three strategies for MLUL. MLUL is relatively insensitive to the parameters changing, hence using various $p$ and $k$ hardly increase the diversity of MLUL.
\begin{table}  
\centering
\caption{The result of Wilcoxon signed rank test with Bergman-Hommel's correction at the 5\% level for EMLS with different strategies to generate diverse sampled datasets in terms of AUCPR.
}
\label{ta:Results_EMLSpk}
\begin{tabular}{@{}cccc|ccc@{}}
\toprule
\multirow{2}{*}{Base} & \multicolumn{3}{c|}{A=MLSOL} & \multicolumn{3}{c}{A=MLUL} \\ 
 & EA & E$_p$A & E$_k$A & EA & E$_p$A & E$_k$A \\ \midrule
BR  & 0/1 & \textbf{2/0} & 0/1 & 0/0 & 0/0 & 0/0\\
MLkNN  & 0/2 & \textbf{1/0} & \textbf{1/0} & 0/0 & 0/0 & 0/0\\
CLR  & 0/2 & \textbf{1/0} & \textbf{1/0} & \textbf{1/0} & 0/1 & 0/0\\
RAkEL  & 0/2 & \textbf{1/0} & \textbf{1/0} & 0/0 & 0/0 & 0/0 \\
COCOA  & 0/2 & \textbf{1/0} & \textbf{1/0} & 0/0 & 0/1 & \textbf{1/0} \\
ECCRU3  & 0/0 & 0/0 & 0/0  & 0/0 & 0/0 & 0/0 \\
\bottomrule
\end{tabular}
\end{table}

\section{Conclusion}

We presented a local label distribution based measure to assess the local imbalance level of multi-label dataset. Based on the local imbalance concept, we proposed two multi-label sampling methods considering all informative labels, in order to make the multi-label dataset easier to be learned. MLSOL selects difficult seed instances and generates more diverse and well-labeled synthetic instances. MLUL removes harmful instances that are easier and hinder their R$k$NNs. Furthermore, we employed MLSOL and MLUL within a simple ensemble framework, which exploits the random aspects of our approaches during the instance selection and synthetic instance generation. The analysis of the relation between measures and performances of six multi-label learning methods on 13 benchmark multi-label datasets shows the effectiveness of the local label distribution based measure. In addition, experimental results demonstrate the advantage of the proposed methods on the compared multi-label sampling approaches, especially within the ensemble framework due to their ability to produce diverse sampled datasets.


%




\ifCLASSOPTIONcompsoc
  \section*{Acknowledgments}
\else
  \section*{Acknowledgment}
\fi

Bin Liu is supported from the China Scholarship Council (CSC) under the Grant CSC No.201708500095.

\ifCLASSOPTIONcaptionsoff
  \newpage
\fi

\bibliographystyle{IEEEtran} %
\bibliography{alma} 

%

\begin{IEEEbiography}[{\includegraphics[width=1in,height=1.25in,clip,keepaspectratio]{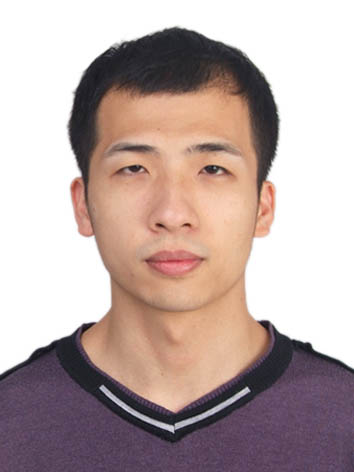}}]{Bin Liu}
received an M.S. degree in computer science from Chongqing University of Posts and Telecommunications, China in 2016. He is currently pursuing a Ph.D. degree in computer science from Aristotle University of Thessaloniki, Greece. His research interests include multi-label learning and class imbalance.
\end{IEEEbiography}

\begin{IEEEbiography}[{\includegraphics[width=1in,height=1.25in,clip]{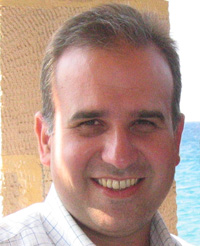}}]{Konstantinos Blekas}
Konstantinos Blekas received the diploma degree (1993) and the Ph.D. degree (1997) in Electrical and Computer Engineering from the National Technical University of Athens. He is currently on the faculty of the Department of Computer Science and Engineering, University of Ioannina, Greece. His research interests include Machine Learning, Reinforcement Learning and Artificial Intelligence.
\end{IEEEbiography}

\begin{IEEEbiography}[{\includegraphics[width=1in,height=1.25in,clip]{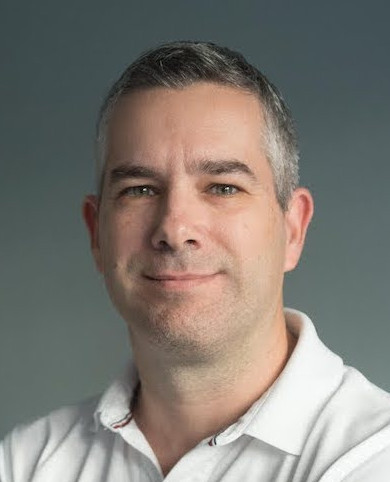}}]{Grigorios Tsoumakas} is an Assistant Professor of Machine Learning and Knowledge Discovery at the School of Informatics of the Aristotle University of Thessaloniki (AUTH) in Greece. He received a degree in Computer Science from AUTH in 1999, an MSc in Artificial Intelligence from the University of Edinburgh, United Kingdom, in 2000 and a PhD in computer science from AUTH in 2005. His research expertise focuses on supervised learning techniques and text mining.
\end{IEEEbiography}






\end{document}